\documentclass[runningheads]{llncs}

\usepackage{eccv}

\usepackage{eccvabbrv}

\usepackage{graphicx}
\usepackage{booktabs}   
\usepackage{multirow}   
\usepackage{makecell}   
\usepackage{xcolor}     
\usepackage{amsmath}    
\usepackage{amssymb}    
\usepackage{colortbl}   
\usepackage{adjustbox}
\usepackage{float}
\usepackage{capt-of}

\definecolor{lightmint}{RGB}{235,250,238}
\definecolor{lightblue}{RGB}{217,234,247}
\definecolor{lightcyan}{RGB}{221,243,250}
\definecolor{lightgreen}{RGB}{228,240,228}
\newcommand{\boldfw}{\fontseries{b}\selectfont} 
\newcommand{\cmark}{\ensuremath{\checkmark}}
\newcommand{\nmark}{--}

\usepackage{hyperref}

\usepackage{orcidlink}

\begin{document}

\title{Hyper3-CLIP: Hierarchy-Conditioned\\Hyperbolic Vision-Language Training}

\titlerunning{Hyper3-CLIP}

\author{Matin Mahmood\inst{1} \and
Antonio Rueda-Toicen\inst{2} \and
Mohamed ElBassat\inst{3} \and
Seifeldin Elkerdany\inst{4} \and
Weixing Wang\inst{2} \and
Gerard de Melo\inst{2}}

\authorrunning{M. Mahmood et al.}

\institute{hyper\textsuperscript{3}labs, Berlin, Germany\\
\email{matin@hyper3labs.com}
\and
Hasso Plattner Institute, University of Potsdam, Potsdam, Germany\\
\email{antonio.rueda-toicen@hpi.de}
\and
Faculty of Computers and Data Science, Alexandria University, Alexandria, Egypt
\and
Faculty of Computer Science and Engineering, Alamein International University, New Alamein, Egypt}

\maketitle
  \begin{abstract}
CLIP-like vision-language models (VLMs) trained with contrastive objectives learn strong global image-text representations, but their Euclidean embeddings and global pooling fail to encode relational structure such as part-whole and parent-child relations. Hyperbolic VLMs address this gap with entailment-based objectives, and text-conditioned variants improve fine-grained alignment through sentence- and phrase-level queries. However, these two lines of work remain separate: hyperbolic VLMs use static image and region features, while query-conditioned methods lack hierarchical geometric structure.
We present Hyper3-CLIP, a hierarchy-conditioned hyperbolic VLM that combines global, local, and global-local contrastive learning with query-conditioned visual pooling. To train the model, we construct lightweight query hierarchies from text, comprising full captions, sentence fragments, localized part descriptions, and extracted phrases. Each query conditions the pooling of visual patches, and the resulting representations support image-text, whole-part, and parent-child entailment losses. Query-conditioned pooling is active only during training.
Hyper3-CLIP improves R@5 and R@10 retrieval on COCO and Flickr, as well as multi-label classification on VOC and COCO, while remaining competitive on hierarchy metrics. We also audit zero-shot prompt sensitivity under fixed prompt regimes and study the effect of the localized GRIT part budget used during training.
Code is available at \url{https://github.com/Hyper3Labs/hyper3-clip}.

\end{abstract}

\section{Introduction}

Vision-language models (VLMs) have become the dominant paradigm for learning transferable multimodal representations. Large-scale contrastive pretraining methods such as CLIP~\cite{radford2021learning}, ALIGN~\cite{jia2021scaling}, LiT~\cite{zhai2022lit}, BLIP~\cite{li2022blip}, and SigLIP~\cite{zhai2023sigmoid} align images and text in a shared embedding space, enabling strong zero-shot recognition, image-text retrieval, and multimodal reasoning.

Despite their success, most VLMs learn representations in Euclidean space using a single global embedding for each image and caption. While effective for instance-level alignment, this formulation provides limited structure for modeling the hierarchical relationships naturally present in visual scenes and language. Images consist of objects and parts, while captions contain sentences, phrases, and localized descriptions that exist at multiple semantic levels.

Hyperbolic geometry provides a natural representation space for hierarchical data, motivating a broad line of work on hyperbolic representation learning~\cite{nickel2017poincare,ganea2018hyperbolic,nickel2018learning,law2019lorentzian,he2025position} and, more recently, hyperbolic vision-language models. MERU~\cite{desai2023hyperbolic} first demonstrated the benefits of hyperbolic embeddings through entailment-based supervision. ATMG~\cite{ramasinghe2024accept} later showed that proximity-based contrastive objectives can hinder hierarchical structure learning in hyperbolic spaces and proposed an angle-based alternative to better preserve geometric hierarchy. HyCoCLIP~\cite{pal2024compositional} incorporated grounded image-box and text-box pairs for compositional alignment, while UNCHA~\cite{kim2026uncha} introduced uncertainty-aware hierarchical alignment for part-to-whole reasoning. More recently, PHyCLIP~\cite{yoshikawa2026phyclip} models hierarchy within individual hyperbolic spaces and compositional semantics via an $\ell_1$-product over multiple hyperbolic factors. Although these methods capture hierarchical semantics, they remain constrained to static image and region representations and do not exploit query-conditioned visual features that dynamically adapt to different textual granularities.

In a separate line of work, recent vision-language models have moved beyond global image-text alignment toward grounded and fine-grained supervision. Methods such as ALBEF~\cite{li2021align}, RegionCLIP~\cite{zhong2022regionclip}, GLIP~\cite{li2022grounded}, Grounding DINO~\cite{liu2024grounding}, BLIP-2~\cite{li2023blip}, and Kosmos-2~\cite{peng2023kosmos} incorporate region-level grounding, open-vocabulary localization, or query-driven visual representations, enabling stronger fine-grained alignment between language and visual content. In particular, $\beta$-CLIP~\cite{zohra2025betaclip} introduces sentence- and phrase-level queries with query-conditioned attention, producing multiple semantically focused visual embeddings from a single image. While effective for improving granularity, these methods do not explicitly incorporate hierarchical geometric structure for modeling part-whole or multi-granular semantic relations, which motivates the use of hyperbolic representations.

To bridge this gap, we propose \textbf{Hyper3-CLIP}\footnote{Pretrained model: \url{https://huggingface.co/hyper3labs/hyper3-clip}}, a hierarchy-conditioned hyperbolic vision-language model that combines query-conditioned visual pooling with hyperbolic hierarchical representation learning. Hyper3-CLIP constructs lightweight textual hierarchies from captions, sentence fragments, localized descriptions, and extracted phrases; each textual query attends to image patch tokens to produce a query-specific visual embedding, which is projected into hyperbolic space alongside its corresponding text representation. The resulting query-level representations are integrated with grounded image-box supervision through hyperbolic contrastive and entailment learning, enabling explicit modeling of both semantic hierarchy and fine-grained visual composition. Through extensive experiments on image-text retrieval benchmarks, we demonstrate that combining hyperbolic hierarchical representations with query-conditioned visual alignment improves image-text retrieval and hierarchy-sensitive image-caption alignment.

Our contribution is the integration of these two lines of work rather than a new objective family or pooling mechanism. Query decomposition and text-conditioned attention pooling follow $\beta$-CLIP~\cite{zohra2025betaclip}, and the base contrastive and grounded box-level entailment objective is inherited from UNCHA~\cite{kim2026uncha}. What is new is their parent-linked hyperbolic combination: each query carries a parent link and instantiates its own visual node in the shared hyperbolic space, and these nodes are supervised through three query-level entailment relations. Query-conditioned pooling is active only during training; at inference, Hyper3-CLIP is a standard dual encoder, and the pooling module adds no inference-time computation.

\section{Method}

\begin{figure}[t]
\centering
\includegraphics[width=\textwidth]{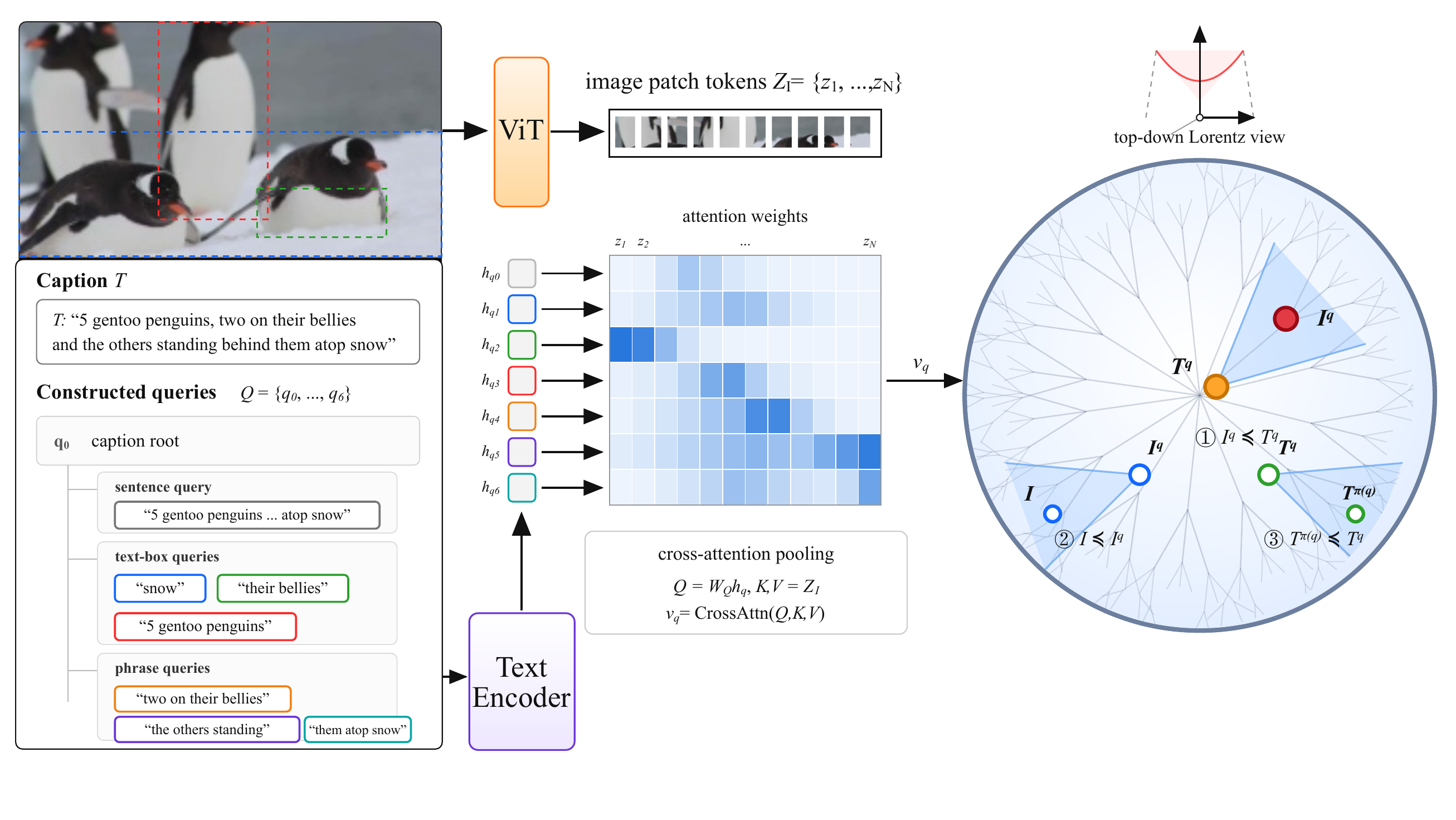}
\caption{\textbf{Hyper3-CLIP overview.}
The method assumes an image along with a whole-image caption and localized image boxes with corresponding text boxes. We define caption, sentence, text-box, and phrase query nodes.
Each text query attends over full-image ViT patch tokens to form a query-conditioned visual node for the hierarchy losses, while the stored image boxes and text boxes remain available for grounded local alignment.}
\label{fig:overview}
\end{figure}

Hyper3-CLIP extends a grounded hyperbolic training setup for image-caption and image-box/text-box pairs.
Each training sample consists of a full image-caption pair $(I,T)$ and grounded box-level pairs $\{(I^\text{box}_j,T^\text{box}_j)\}_{j=1}^{P}$, where $I^\text{box}_j$ is an image box (a localized crop) and $T^\text{box}_j$ is the corresponding text box, typically a noun or phrase from the caption.
The inherited objective supplies the contrastive and entailment losses over these full-image and localized nodes.
Hyper3-CLIP maintains this base structure and adds query-conditioned visual nodes derived from the full-image patch tokens.
Appendix~\ref{app:base-objective} gives the Lorentz geometry and inherited HyCoCLIP/UNCHA loss terms.
Relative to this inherited base, Hyper3-CLIP adds three components: the hierarchical query construction (Sec.~\ref{sec:query-construction}), the query-conditioned visual pooling module adapted from $\beta$-CLIP~\cite{zohra2025betaclip} (Sec.~\ref{sec:query-pooling}), and the query-level entailment relations added to the training objective (Sec.~\ref{sec:training-objective}).

\subsection{Hierarchical Query Construction}
\label{sec:query-construction}
For each training sample, Hyper3-CLIP builds an additional query set $\mathcal{Q}$ from the sample text annotations: the caption and localized text boxes. The caption is the root query. Sentence queries are sentence-level fragments of the caption; text-box queries reuse the localized text boxes; phrase queries are lightweight phrases extracted from the caption. Unlike text-box queries, phrase queries have no image-box annotation.

Each query stores a parent $\pi(q)$ and loss weight $w(q)$. Table~\ref{tab:query-construction} summarizes the query types and parent links.
The weights encode a simple reliability prior: sentence fragments are direct caption spans and receive full weight, text boxes are grounded but shorter and less contextual, extracted phrases are the weakest textual units, and the caption root has zero query weight because the full image-caption pair is already supervised by the inherited global objective.

\begin{center}
\begin{minipage}{0.76\linewidth}
\captionof{table}{Query types used to build the hierarchy. The caption root provides parent context; sentence, text-box, and phrase queries receive nonzero entailment weights.}
\label{tab:query-construction}

\small
\setlength{\tabcolsep}{5.0pt}
\renewcommand{\arraystretch}{1.08}
\begin{adjustbox}{max width=\linewidth}
\begin{tabular}{@{}lll@{}}
\toprule
Query & Source and parent & $w(q)$ \\
\midrule
Caption root & caption; none & $0.0$ \\
Sentence & caption sentence; caption & $1.0$ \\
Text box & text box; caption & $0.75$ \\
Phrase & caption phrase; sentence/caption & $0.5$ \\
\bottomrule
\end{tabular}
\end{adjustbox}
\end{minipage}
\end{center}

Text boxes continue to supervise their paired image boxes through the inherited objective. Hyper3-CLIP also uses each text box to condition pooling over full-image patch tokens.

\subsection{Query-Conditioned Visual Pooling}
\label{sec:query-pooling}
Query-conditioned visual pooling converts each text query into a separate visual node for its owner image.

Let $Z_I \in \mathbb{R}^{N \times d_v}$ denote the full-image patch tokens produced by the image encoder after removing the class token, and let $h_q \in \mathbb{R}^{d_t}$ be the text-encoder representation of query $q$.
A learned linear map $W_Q \in \mathbb{R}^{d_v \times d_t}$ projects $h_q$ to the visual-token dimension, after which it attends over the image patch tokens:
\[
\widetilde{v}_q =
\operatorname{MHA}\!\left(Q=W_Qh_q,\;K=Z_I,\;V=Z_I\right),
\qquad
v_q = \widetilde{v}_q +
\operatorname{MLP}\!\left(\operatorname{LN}(\widetilde{v}_q)\right).
\]
The cross-attention block uses 8 heads, and the MLP has hidden dimension $4d_v$ with a GELU activation.
This text-conditioned attention pooling follows $\beta$-CLIP~\cite{zohra2025betaclip}; Hyper3-CLIP uses it to instantiate hyperbolic query nodes rather than Euclidean embeddings.
The resulting representation $v_q$ is the visual node for query $q$.
It is pooled from the full image, while the text query guides the pooling.
Finally, $v_q$ passes through the shared image projection and Lorentz lift, while $h_q$ passes through the shared text projection and lift; we denote the resulting hyperbolic embeddings by $I^q$ and $T^q$, respectively.
Query-conditioned pooling is used only during training, and evaluation uses the standard global embeddings of the dual encoder.

\subsection{Training Objective}
\label{sec:training-objective}
The training loss combines contrastive alignment with hyperbolic entailment and uncertainty calibration.
The contrastive term $\mathcal{L}^{\mathrm{con}}_{\mathrm{UNCHA}}$ keeps the UNCHA global, local, and global-local alignments.
The base entailment term $\mathcal{L}^{\mathrm{ent}}_{\mathrm{UNCHA}}$ keeps the inherited relations among $I$, $T$, and the grounded box-level pairs $\{(I^{box}_j,T^{box}_j)\}_{j=1}^{P}$.

For each non-root query $q \in \mathcal{Q}$ with $w(q)>0$, the query-conditioned nodes add three entailment relations:
\[
(1)\; I^q \preceq T^q,\qquad
(2)\; I \preceq I^q,\qquad
(3)\; T^{\pi(q)} \preceq T^q .
\]
Following MERU~\cite{desai2023hyperbolic}, $E(a \preceq b)$ denotes the hyperbolic entailment-cone violation for an ordered pair.
Node $b$ roots a cone of more specific points, and node $a$ is penalized when it lies outside that cone.
Relation (1) is the query-level analogue of image-to-text entailment.
For relation (2), the ordering is defined by semantic specificity rather than by the computational origin of the representations.
Query-conditioned pooling removes image content unrelated to the query, so $I^q$ represents a broader query-specific concept, whereas $I$ retains the complete scene and its additional context.
Thus, we treat the full image as more specific than the query-conditioned visual node, consistent with the inherited relation $I \preceq I^{box}$, where the complete scene is more specific than a localized part.
Relation (3) preserves the query hierarchy on the text side: a parent query is treated as more specific than its child because it contains additional contextual information.

The query terms are summed over non-root queries with reliability weights $w(q)$ and calibrated with the same uncertainty mechanism as the inherited UNCHA objective to form $\mathcal{L}^{\mathrm{ent}}_{\mathrm{query}}$.
The full objective is
\[
\mathcal{L}
=
\mathcal{L}^{\mathrm{con}}_{\mathrm{UNCHA}}
+ \lambda_{\mathrm{ent}}
\left(
\mathcal{L}^{\mathrm{ent}}_{\mathrm{UNCHA}}
+ \mathcal{L}^{\mathrm{ent}}_{\mathrm{query}}
\right).
\]
\section{Experiments}

\smallskip
\noindent\textbf{GRIT data.} Hyper3-CLIP is trained on GRIT~\cite{peng2023kosmos} using all available localized parts up to a cap of 5 parts per image.
Figure~\ref{fig:max-parts-sweep} studies this max-parts setting and shows that it retains 99.38\% of localized part instances in the processed training data.

\smallskip
\noindent\textbf{Query construction.} Queries are constructed by rule-based string processing, without a learned parser.
For each image, candidates are added in a fixed order: the caption root, then sentence queries, then text-box queries, then extracted phrases.
Every candidate is normalized by collapsing whitespace runs to single spaces and is accepted only if it is at least 3 characters long and its case-insensitive text has not already been accepted for that image.
Sentence queries split the caption at whitespace following ``.'', ``!'', ``?'', or ``;'' and at line breaks, keeping the first 5 candidates; their parent is the caption.
Text-box queries reuse the localized text boxes, also with the caption as parent.
Phrase extraction splits the caption at commas, semicolons, colons, and parentheses, and at the connector words \textit{and}, \textit{with}, \textit{near}, \textit{beside}, \textit{behind}, \textit{under}, \textit{above}, \textit{around}, and \textit{next to} (matched case-insensitively); each resulting fragment is reduced to its alphanumeric words, allowing internal hyphens and apostrophes and discarding other punctuation.
A fragment of 2--8 words yields one phrase; a longer fragment is cut into windows of at most 6 words starting every 4 words, keeping windows with at least 2 words; fragments with fewer than 2 words are discarded.
Text-box and phrase queries share a budget of 30 accepted queries per image, filled by text boxes first; each accepted phrase takes the first accepted sentence query as parent, or the caption when no sentence query exists.
Finally, at most 6 queries per image are kept, truncating in construction order.

\smallskip
\noindent\textbf{Architecture.} Following prior hyperbolic VLMs~\cite{pal2024compositional,kim2026uncha,yoshikawa2026phyclip}, we use a CLIP-style dual-encoder architecture with a ViT image encoder and a Transformer text encoder.
The main Hyper3-CLIP result uses a ViT-B/16 image backbone, the CLIP-B/32 text-encoder architecture, a 512-dimensional multimodal embedding, Lorentz projection, and random initialization for both towers.
ViT-S/16 appears in the published baselines and controlled ablations, but not as a main Hyper3-CLIP result.

\smallskip
\noindent\textbf{Evaluation.} All downstream results are zero-shot evaluations of frozen checkpoints.
The cross-attention pooling module is dropped at inference: every evaluation scores images and text with the standard global dual-encoder embeddings, so the pooling module adds no computation at inference time.
We report image-text retrieval, ImageNet hierarchy metrics, 16-dataset classification, and VOC/COCO multi-label classification.

\subsection{Zero-Shot Retrieval and Hierarchy}

In zero-shot retrieval, the model must find the matching caption for an image, and vice versa, using only its learned multimodal alignment. Using COCO~\cite{lin2014microsoft} and Flickr30K~\cite{plummer2015flickr30k}, we measure Recall@K for image-to-text and text-to-image retrieval.
We also report ImageNet hierarchy metrics to test whether the hyperbolic embedding preserves class-label structure.

\begin{table}[t]
\centering
\scriptsize
\setlength{\tabcolsep}{3.0pt}
\renewcommand{\arraystretch}{1.08}
\caption{Zero-shot image-text retrieval and ImageNet hierarchy evaluation for the ViT-B/16 setting. Published baseline values are from UNCHA Table~2~\cite{kim2026uncha}; Hyper3-CLIP is evaluated with the same retrieval and hierarchy metrics. Paired retrieval columns report R@5/R@10. T and I denote text retrieval and image retrieval. Higher is better except TIE and LCA. Bold marks the best value in each column, with ties bolded.}
\begin{adjustbox}{max width=\textwidth}
\begin{tabular}{llrrrrrrrrrrrrr}
\toprule
Backbone & Model & \multicolumn{4}{c}{\cellcolor{lightblue}\textbf{Text retrieval}} & \multicolumn{4}{c}{\cellcolor{lightcyan}\textbf{Image retrieval}} & \multicolumn{5}{c}{\cellcolor{lightgreen}\textbf{Hierarchy}} \\
\cmidrule(lr){3-6} \cmidrule(lr){7-10} \cmidrule(lr){11-15}
& & \multicolumn{2}{c}{COCO} & \multicolumn{2}{c}{Flickr} & \multicolumn{2}{c}{COCO} & \multicolumn{2}{c}{Flickr} & TIE$\downarrow$ & LCA$\downarrow$ & J$\uparrow$ & H-P$\uparrow$ & H-R$\uparrow$ \\
\cmidrule(lr){3-4} \cmidrule(lr){5-6} \cmidrule(lr){7-8} \cmidrule(lr){9-10}
& & R@5 & R@10 & R@5 & R@10 & R@5 & R@10 & R@5 & R@10 & & & & & \\
\multirow{6}{*}{ViT-B/16}
& CLIP & 71.4 & 81.5 & 93.6 & 96.9 & 57.4 & 68.5 & 83.5 & 89.9 & 3.60 & 2.21 & 0.79 & 0.85 & 0.85 \\
& MERU & 72.3 & 82.0 & 93.5 & 96.2 & 57.4 & 68.6 & 84.0 & 90.0 & 3.63 & 2.22 & 0.78 & 0.85 & 0.85 \\
& ATMG & 62.9 & 74.0 & 85.1 & 92.2 & 51.2 & 62.6 & 78.0 & 85.3 & 4.19 & 2.48 & 0.75 & 0.83 & 0.83 \\
& HyCoCLIP & 72.0 & 82.0 & 92.6 & 95.4 & 58.4 & 69.3 & 84.9 & 90.3 & 3.17 & 2.05 & 0.81 & 0.87 & 0.87 \\
& UNCHA & 72.7 & 82.7 & 91.4 & 95.9 & 60.0 & 71.0 & 84.9 & 91.2 & \boldfw{2.94} & \boldfw{1.96} & \boldfw{0.83} & \boldfw{0.88} & \boldfw{0.88} \\
& \cellcolor{gray!10}Hyper3-CLIP & \cellcolor{gray!10}\boldfw{75.7} & \cellcolor{gray!10}\boldfw{84.3} & \cellcolor{gray!10}\boldfw{95.4} & \cellcolor{gray!10}\boldfw{97.6} & \cellcolor{gray!10}\boldfw{63.0} & \cellcolor{gray!10}\boldfw{73.2} & \cellcolor{gray!10}\boldfw{86.4} & \cellcolor{gray!10}\boldfw{91.4} & \cellcolor{gray!10}3.16 & \cellcolor{gray!10}2.08 & \cellcolor{gray!10}0.82 & \cellcolor{gray!10}0.87 & \cellcolor{gray!10}\boldfw{0.88} \\
\bottomrule
\end{tabular}
\end{adjustbox}
\label{tab:main-results}
\end{table}

The results are given in Table~\ref{tab:main-results}.
Hyper3-CLIP improves over UNCHA across all eight considered COCO/Flickr retrieval settings.
On hierarchy metrics, it does not improve TIE or LCA, but remains competitive with UNCHA.

\subsection{Zero-Shot Classification}

In zero-shot classification, the model assigns labels to images from categories without using dataset-specific training data.
This evaluates how well the learned visual-semantic embeddings generalize to broad, unseen concepts. We measure mean-per-class accuracy on a standard 16-dataset zero-shot classification benchmark that includes ImageNet~\cite{deng2009imagenet}.

\begin{table*}[t]
\centering
\scriptsize
\setlength{\tabcolsep}{3.2pt}
\renewcommand{\arraystretch}{1.08}
\caption{Zero-shot classification on the 16-dataset CLIP evaluation suite. Values are mean-per-class accuracy. Food-101, CUB, and Flowers, marked with $^\ast$, use the common Photo prompt audited in Table~\ref{tab:prompt-sensitivity}; the remaining datasets use the baseline evaluation prompts. Bold marks the best value in each column.}
\begin{adjustbox}{max width=\textwidth}
\begin{tabular}{lrrrrrrrrrrrrrrrrr}
\toprule
& \multicolumn{6}{c}{\cellcolor{lightblue}\textbf{General}} & \multicolumn{6}{c}{\cellcolor{lightcyan}\textbf{Fine-grained}} & \multicolumn{4}{c}{\cellcolor{lightgreen}\textbf{Misc.}} & \\
\cmidrule(lr){2-7} \cmidrule(lr){8-13} \cmidrule(lr){14-17}
Model & \rotatebox{90}{IN} & \rotatebox{90}{C10} & \rotatebox{90}{C100} & \rotatebox{90}{SUN} & \rotatebox{90}{Cal.} & \rotatebox{90}{STL} & \rotatebox{90}{Food$^\ast$} & \rotatebox{90}{CUB$^\ast$} & \rotatebox{90}{Cars} & \rotatebox{90}{Air.} & \rotatebox{90}{Pets} & \rotatebox{90}{Flwr.$^\ast$} & \rotatebox{90}{DTD} & \rotatebox{90}{Euro.} & \rotatebox{90}{RES.} & \rotatebox{90}{C211} & Avg. \\
\midrule
HyCoCLIP-B/16 & 45.80 & 88.80 & 60.10 & 57.20 & 81.30 & 95.00 & 61.14 & 17.76 & 11.60 & 3.70 & 56.80 & 28.79 & 29.40 & 35.80 & 45.60 & \boldfw{6.50} & 45.33 \\
UNCHA-B/16 & \boldfw{48.80} & 90.40 & 63.20 & 57.70 & \boldfw{83.90} & 95.70 & 64.10 & 18.74 & 14.00 & 3.80 & 57.10 & 30.75 & \boldfw{30.30} & \boldfw{41.30} & \boldfw{52.70} & 6.10 & 47.41 \\
PHyCLIP-B/16 & 44.31 & 89.33 & 59.05 & 55.32 & 76.35 & 94.84 & 62.40 & 17.97 & 10.89 & 3.24 & 54.18 & 28.27 & 25.50 & 36.29 & 48.22 & 5.56 & 44.48 \\
\cellcolor{gray!10}Hyper3-CLIP-B/16 & \cellcolor{gray!10}46.98 & \cellcolor{gray!10}\boldfw{91.14} & \cellcolor{gray!10}\boldfw{65.64} & \cellcolor{gray!10}\boldfw{59.39} & \cellcolor{gray!10}83.42 & \cellcolor{gray!10}\boldfw{96.56} & \cellcolor{gray!10}\boldfw{67.31} & \cellcolor{gray!10}\boldfw{19.36} & \cellcolor{gray!10}\boldfw{16.12} & \cellcolor{gray!10}\boldfw{4.33} & \cellcolor{gray!10}\boldfw{62.57} & \cellcolor{gray!10}\boldfw{39.66} & \cellcolor{gray!10}29.41 & \cellcolor{gray!10}28.52 & \cellcolor{gray!10}48.94 & \cellcolor{gray!10}6.06 & \cellcolor{gray!10}\boldfw{47.84} \\
\bottomrule
\end{tabular}
\end{adjustbox}
\label{tab:zero-shot-16}
\end{table*}

Under the stated prompt protocol, Hyper3-CLIP obtains the best 16-dataset average, reaching 47.84 compared with 47.41 for UNCHA.
The gains are concentrated on CIFAR-10, CIFAR-100, SUN397, STL-10, Food-101, Flowers, Cars, Aircraft, and Pets, whereas weaknesses remain for Country211, DTD, and EuroSAT.
Section~\ref{sec:prompt-sensitivity} audits the three datasets evaluated with the common Photo prompt and reports their corresponding official-prompt results separately.

\subsection{Zero-Shot Multi-Label Classification}

For VOC and COCO multi-label classification, each image can contain multiple valid object categories.
We score each class independently and report mean Average Precision (mAP), following the framing used by UNCHA.

\begin{table}[t]
\centering
\scriptsize
\setlength{\tabcolsep}{5.0pt}
\renewcommand{\arraystretch}{1.08}
\caption{Zero-shot multi-label classification on VOC and COCO under a single evaluator. Values are mAP; Avg. is the mean of VOC and COCO. Higher is better, and bold marks the best value in each column.}
\begin{adjustbox}{max width=\textwidth}
\begin{tabular}{llrrr}
\toprule
Backbone & Model & \multicolumn{3}{c}{\cellcolor{lightcyan}\textbf{Multi-label classification}} \\
\cmidrule(lr){3-5}
& & VOC & COCO & Avg. \\
\midrule
\multirow{6}{*}{ViT-B/16}
& CLIP & 68.55 & 44.89 & 56.72 \\
& MERU & 71.29 & 45.33 & 58.31 \\
& HyCoCLIP & 74.89 & 50.73 & 62.81 \\
& UNCHA & 75.51 & 50.93 & 63.22 \\
& PHyCLIP & 73.29 & 49.64 & 61.47 \\
& \cellcolor{gray!10}Hyper3-CLIP & \cellcolor{gray!10}\boldfw{78.20} & \cellcolor{gray!10}\boldfw{54.28} & \cellcolor{gray!10}\boldfw{66.24} \\
\bottomrule
\end{tabular}
\end{adjustbox}
\label{tab:voc-coco}
\end{table}

Under a shared evaluator, Hyper3-CLIP improves over UNCHA by 2.69 mAP on VOC and 3.35 mAP on COCO.
Because the VOC/COCO values reported in UNCHA Table~5 use a different protocol, we do not mix those numbers into the main comparison.

\section{Ablation Study}
\label{sec:ablation}

We further conduct a series of analyses, including ablating the query-conditioned objective to determine which relations are responsible for the hierarchy-sensitive gains and whether those gains preserve standard retrieval performance.

\subsection{Query-Conditioned Objective Components}

The full objective augments the grounded contrastive loss with three query-level entailment relations.
These relations align the query-conditioned visual node with its query text, $I^q{\preceq}T^q$; treat the full-image node as more specific than the query-conditioned visual node, $I{\preceq}I^q$; and treat the construction-parent text node as more specific than its child query, $T^{\pi(q)}{\preceq}T^q$.
To isolate their effects, each variant is first trained with only the base objective for 80k steps, then continues training to 100k steps with the query entailment terms specified in Table~\ref{tab:objective-components}.
Data, batch size, optimizer schedule, and seed are fixed; the control rows keep all query entailment terms enabled and modify only the named control.

We evaluate retrieval and hierarchy-sensitive image-caption alignment.
For hierarchy evaluation, following the image-caption structure of HierarCaps~\cite{alper2024hierarcaps}, positives pair an image with captions from its hierarchy and negatives pair the image with captions sampled from other fine-grained nodes.
Pairs are ranked by the model's hyperbolic entailment value, $p(a{\preceq}b)=\max(1-2\phi(a,b)/\pi,0)$, where $\phi$ is the exterior cone angle between the image and text embeddings.
We report Average Precision (AP) and AUROC over 4,000 positive and 100,000 negative pairs from 1,000 images.
Avg. R@10 is the mean of the four COCO/Flickr text- and image-retrieval R@10 values.

\begin{table}[!htbp]
\centering
\caption{Component ablation for the query-conditioned objective. A checkmark indicates that the corresponding query loss from Sec.~\ref{sec:training-objective} is enabled. The control rows keep all three query losses enabled while replacing text-conditioned visual pooling, reversing the direction of the visual whole-to-query relation, or corrupting text-parent links. $\Delta$AP and $\Delta$R@10 are measured relative to the no-query baseline.}
\setlength{\tabcolsep}{2.4pt}
\scriptsize
\begin{adjustbox}{max width=\textwidth}
\begin{tabular}{lcccrrrr}
\toprule
Variant & $I^q{\preceq}T^q$ & $I{\preceq}I^q$ & $T^{\pi(q)}{\preceq}T^q$ & Hierarchy AP & $\Delta$AP & AUROC & $\Delta$R@10 \\
\midrule
No query losses & \nmark & \nmark & \nmark & 76.21 & 0.00 & 98.34 & 0.00 \\
Query image-text only & \cmark & \nmark & \nmark & 75.34 & $-$0.87 & 98.24 & $-$0.46 \\
Visual hierarchy only & \nmark & \cmark & \nmark & 71.85 & $-$4.36 & 98.01 & $-$0.77 \\
Text hierarchy only & \nmark & \nmark & \cmark & 78.86 & +2.65 & 98.44 & $-$0.09 \\
\cellcolor{gray!10}Full objective & \cellcolor{gray!10}\cmark & \cellcolor{gray!10}\cmark & \cellcolor{gray!10}\cmark & \cellcolor{gray!10}\boldfw{79.85} & \cellcolor{gray!10}\boldfw{+3.64} & \cellcolor{gray!10}\boldfw{98.55} & \cellcolor{gray!10}+0.06 \\
\midrule
Mean-pooled visual nodes & \cmark & \cmark & \cmark & 75.97 & $-$0.25 & 98.32 & \boldfw{+0.15} \\
Reversed visual order ($I^q{\preceq}I$) & \cmark & \cmark & \cmark & 76.06 & $-$0.15 & 98.29 & $-$0.09 \\
Shuffled text parents & \cmark & \cmark & \cmark & 77.22 & +1.00 & 98.40 & $-$0.93 \\
\bottomrule
\end{tabular}
\end{adjustbox}
\label{tab:objective-components}
\end{table}

In the results in Table~\ref{tab:objective-components}, we observe that the full objective achieves the highest caption-hierarchy AP, improving over the no-query baseline by 3.64 points while leaving retrieval essentially unchanged.
Among the single-relation variants, the text parent-to-query term is the strongest and recovers 73\% of the full AP gain ($2.65/3.64$).
The query image-text and visual whole-to-query terms do not improve AP alone, but the full objective improves over the text-only variant by 0.99 AP; their effect is therefore visible in the jointly trained objective rather than as separate single-term gains.

The controls further support the role of the proposed hierarchy construction.
Replacing text-conditioned visual pooling with mean-pooled visual nodes removes the AP gain.
Reversing the visual hierarchy eliminates the full objective's hierarchy gain: AP falls from 79.85 to 76.06, slightly below the no-query baseline of 76.21, while average retrieval remains near baseline.
This matched control confirms that the hierarchy improvement under the Table~\ref{tab:objective-components} protocol depends on the proposed $I{\preceq}I^q$ direction rather than merely adding a visual-hierarchy constraint.
Shuffling the text-parent links substantially lowers AP by 2.63 points relative to the full objective.
Together, these results show that query-conditioned training improves hierarchy-sensitive image-caption alignment while preserving retrieval performance.

\subsection{Prompt Sensitivity and Zero-Shot Protocol}
\label{sec:prompt-sensitivity}

The zero-shot classification results in Table~\ref{tab:zero-shot-16} are most sensitive to prompt choice on Food-101, CUB, and Flowers102.
We therefore evaluate these three datasets under both the \textit{Official} prompt template and a shared \textit{Photo} prompt template for every model.
The exact prompt strings are:
\begin{center}
\scriptsize
\setlength{\tabcolsep}{4pt}
\renewcommand{\arraystretch}{1.08}
\begin{adjustbox}{max width=0.96\linewidth}
\begin{tabular}{@{}lp{0.72\linewidth}@{}}
\toprule
Prompt regime & Templates \\
\midrule
Official Food-101 & \texttt{food : \{\}.}; \texttt{food porn : \{\}.} \\
Official CUB & \texttt{bird pics : \{\}.}; \texttt{birding : \{\}.}; \texttt{birds : \{\}.}; \texttt{bird photography : \{\}.} \\
Official Flowers102 & \texttt{flowers : \{\}.} \\
Photo & \texttt{a photo of a \{\}.} \\
\bottomrule
\end{tabular}
\end{adjustbox}
\end{center}
The remaining 13 datasets in the 16-dataset suite are unchanged across the reported averages.

\begin{table}[!htbp]
\centering
\caption{Prompt sensitivity on the three zero-shot classification datasets whose inherited prompts were unstable. The Photo prompt template uses the same CLIP-style prompt family for every model and every dataset in this audit. The 16-dataset average changes only the three prompt-sensitive columns; all other zero-shot columns are held fixed. Values are mean-per-class accuracy, and bold marks the best value in each column.}
\setlength{\tabcolsep}{4.0pt}
\small
\begin{adjustbox}{max width=\textwidth}
\begin{tabular}{lrrrrrr}
\toprule
& \multicolumn{3}{c}{Food/CUB/Flowers avg.} & \multicolumn{3}{c}{16-dataset avg.} \\
\cmidrule(lr){2-4} \cmidrule(lr){5-7}
Model & Official & Photo prompt & $\Delta$ & Official & Photo prompt & $\Delta$ \\
\midrule
HyCoCLIP-B/16 & 33.09 & 35.90 & +2.81 & 44.82 & 45.33 & +0.51 \\
UNCHA-B/16 & 33.90 & 37.86 & +3.96 & 46.69 & 47.41 & +0.72 \\
\cellcolor{gray!10}Hyper3-CLIP-B/16 & \cellcolor{gray!10}3.26 & \cellcolor{gray!10}\boldfw{42.11} & \cellcolor{gray!10}\boldfw{+38.85} & \cellcolor{gray!10}40.55 & \cellcolor{gray!10}\boldfw{47.84} & \cellcolor{gray!10}\boldfw{+7.29} \\
\bottomrule
\end{tabular}
\end{adjustbox}
\label{tab:prompt-sensitivity}
\end{table}

The common Photo prompt template improves the recorded HyCoCLIP and UNCHA three-dataset averages relative to the official template, showing that these columns are prompt-sensitive even for released baselines.
Hyper3-CLIP performance collapses on the Official prompt template for these datasets, but it recovers when the Photo prompt template is used.
This suggests that short query phrases such as \texttt{flower} can interact poorly with the short Official prompt prefixes used for these datasets, possibly because training supervises the text encoder with compositional multi-word queries rather than terse class-name templates.

\subsection{Parts per Image}

\begin{figure}[!htbp]
\centering
\vspace{-0.4em}
\includegraphics[width=0.95\linewidth]{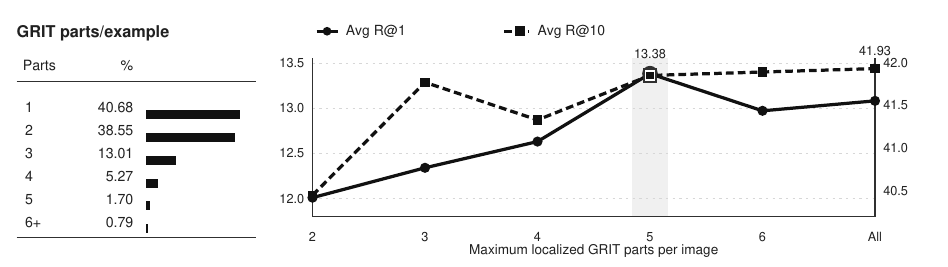}
\vspace{-0.6em}
\caption{Left: GRIT parts per example. Right: 10k proxy sweep over the cap on localized parts per image. Solid circles: Avg R@1; dashed squares: Avg R@10.}
\label{fig:max-parts-sweep}
\vspace{-0.4em}
\end{figure}

The processed GRIT data exhibits a long tail in the number of localized parts per image.
The left panel of Fig.~\ref{fig:max-parts-sweep} summarizes an exact scan of 2,051 shards with 13,149,251 examples and 25,185,017 localized parts.
The mean is 1.9153 parts per example, and the maximum observed example has 20 parts.
Although 40.68\% of examples have exactly one part and 79.23\% have only one or two parts, the tail introduces a small number of examples with many local boxes.

This tail can also affect HyCoCLIP, which trains on localized GRIT boxes.
It is more consequential for UNCHA and Hyper3-CLIP training because each retained part can contribute not only a local image-text correspondence, but also part-whole, uncertainty-weighted, and query-conditioned hierarchy terms.
If the additional boxes on long-tail examples are marginal or noisy, retaining more localized parts per image can therefore add noisy constraints while also increasing the number of local embeddings and loss terms contributed by a single example.
We therefore sweep caps of 2 to 6 localized parts per image, plus an unbounded all-parts condition, and report the 10k proxy retrieval results in the right panel of Fig.~\ref{fig:max-parts-sweep}.

Five parts gives the best average R@1 while retaining 99.38\% of localized part instances and leaving 99.20\% of examples untruncated.
Using all parts marginally improves Avg R@10 but does not improve Avg R@1.
This suggests that the remaining long-tail parts add limited retrieval benefit while increasing the number of per-example embeddings and hierarchy losses.

\section{Conclusion}

We presented Hyper3-CLIP, a hyperbolic vision-language model that introduces query-conditioned visual nodes into grounded image-text training.
From each caption, Hyper3-CLIP constructs a lightweight query hierarchy spanning the full caption, sentence fragments, localized text boxes, and extracted phrases.
Each query pools full-image patch tokens through query-conditioned cross-attention, producing a visual node that is embedded in the shared hyperbolic space alongside its query text.
These nodes support three hierarchy-aware entailment relations: between each query's visual node and its query text, between the whole image and its query nodes, and between parent and child texts.
Standard global, local, and global-local contrastive alignment is retained, and since the pooling module is used only during training, inference keeps the standard dual-encoder cost of CLIP.

Zero-shot evaluations support this design.
In the ViT-B setting, Hyper3-CLIP improves R@5 and R@10 retrieval on COCO and Flickr while remaining competitive on ImageNet hierarchy metrics. Under the stated prompt and evaluator protocols, it also records the highest reported 16-dataset zero-shot classification average and VOC/COCO multi-label mAP.
Controlled ablations show that the full query-conditioned objective raises caption-hierarchy entailment AP without reducing average retrieval. They also show that correct text-parent links and query-conditioned visual pooling are important for this improvement.
The prompt-sensitivity and parts-per-image analyses provide two practical diagnostics for this setting: zero-shot classification can depend sharply on prompt wording, and a moderate cap on localized parts can retain nearly all grounded supervision while avoiding long-tail training overhead.

Overall, these results suggest that conditioning visual representations on a textual query hierarchy is an effective way to inject fine-grained, multi-granular structure into hyperbolic vision-language training.
Future work could explore using query-conditioned visual nodes during inference, as well as richer query hierarchies and broader compositional evaluations of hierarchy-aware multimodal models.

\clearpage
\appendix
\renewcommand{\theHsection}{appendix.\arabic{section}}
\renewcommand{\theHsubsection}{appendix.\arabic{section}.\arabic{subsection}}
\section{Base Hyperbolic Objective Details}
\label{app:base-objective}

This appendix gives the geometric and objective background inherited by Hyper3-CLIP.
The main method builds on these components and adds query-conditioned visual nodes.

\subsection{Hyperbolic Space and the Lorentz Model}
Hyperbolic space has exponential volume growth, making it suitable for embeddings with tree-like or part-whole structure.
Following MERU, HyCoCLIP, and UNCHA, the model represents image and text embeddings on the Lorentz manifold.
With curvature magnitude $c>0$, the $d$-dimensional Lorentz model is
\[
\mathbb{L}_c^d
=
\left\{
x=(x_0,\bar{x}) \in \mathbb{R}^{d+1}:
\langle x,x\rangle_L=-1/c,\; x_0>0
\right\},
\]
where the Lorentz inner product is
\[
\langle x,y\rangle_L = -x_0y_0 + \bar{x}^{\top}\bar{y}.
\]
The geodesic distance between two points is
\[
d_{\mathbb{L}}(x,y)
=
\frac{1}{\sqrt{c}}\operatorname{arcosh}\!\left(-c\langle x,y\rangle_L\right).
\]
Image and text encoders produce Euclidean features that are projected to this manifold, and contrastive retrieval uses $-d_{\mathbb{L}}(x,y)$ as the similarity score.

Hyperbolic entailment uses cones rooted at more general nodes.
For an ordered pair $a \preceq b$, the node $a$ is treated as more specific and is penalized if it falls outside the cone of $b$.
Let $E(a \preceq b)$ denote this cone penalty, computed from the exterior angle between $a$ and the cone rooted at $b$.
This notation is used in the method section for both inherited part-whole relations and the added query-conditioned relations.

\subsection{Inherited HyCoCLIP and UNCHA Losses}
Let $g_I(\cdot)$ and $g_T(\cdot)$ denote image and text encoders after projection to the Lorentz manifold.
For a batch, let $\mathcal{L}_c^*(A,B)$ denote the directional contrastive loss that matches embeddings of type $A$ to their paired embeddings of type $B$ and treats other batch elements as negatives.
HyCoCLIP extends global image-caption alignment with grounded image-box and text-box supervision.
In the notation of this paper, the inherited contrastive groups are
\[
\mathcal{L}^{\mathrm{con}}_{\mathrm{global}}
=
\mathcal{L}_c^*(I,T)+\mathcal{L}_c^*(T,I),
\]
\[
\mathcal{L}^{\mathrm{con}}_{\mathrm{local}}
=
\mathcal{L}_c^*(I^{box},T^{box})+\mathcal{L}_c^*(T^{box},I^{box}),
\]
\[
\mathcal{L}^{\mathrm{con}}_{\mathrm{global-local}}
=
\mathcal{L}_c^*(I^{box},T)+\mathcal{L}_c^*(T^{box},I).
\]
UNCHA preserves these groups and modulates the global-local terms using part uncertainty, so parts estimated to be more representative of the full sample receive stronger alignment.

The inherited entailment objective applies $E(a \preceq b)$ to four HyCoCLIP relations:
\[
I \preceq T,\qquad
I^{box} \preceq T^{box},\qquad
I \preceq I^{box},\qquad
T \preceq T^{box}.
\]
The first two relations are cross-modal image-text entailments at global and local granularity.
The last two are part-whole entailments within each modality: the full image or caption contains scene context that is absent from its image box or text box, so the full node is treated as more specific.

UNCHA keeps these relations but introduces uncertainty-guided calibration for the part-whole terms.
Its uncertainty estimate is derived from hyperbolic radius and is used to reduce the influence of ambiguous or weakly representative parts while preserving the same underlying compositional ordering.
The inherited base objective is written compactly as
\[
\mathcal{L}_{\mathrm{UNCHA}}
=
\mathcal{L}^{\mathrm{con}}_{\mathrm{UNCHA}}
+
\lambda_{\mathrm{ent}}
\mathcal{L}^{\mathrm{ent}}_{\mathrm{UNCHA}}.
\]
Hyper3-CLIP retains this base objective, while adding the query-conditioned entailment terms described in Sec.~\ref{sec:training-objective}.

%
%
\bibliographystyle{splncs04}
\bibliography{main}

@String(CVPR  = {IEEE Conf. Comput. Vis. Pattern Recog.})

@String(ICCV  = {Int. Conf. Comput. Vis.})

@String(ECCV  = {Eur. Conf. Comput. Vis.})

@String(CVPR  = {CVPR})

@String(ICCV  = {ICCV})

@String(ECCV  = {ECCV})

@inproceedings{radford2021learning,
  title={Learning transferable visual models from natural language supervision},
  author={Radford, Alec and Kim, Jong Wook and Hallacy, Chris and Ramesh, Aditya and Goh, Gabriel and Agarwal, Sandhini and Sastry, Girish and Askell, Amanda and Mishkin, Pamela and Clark, Jack and Krueger, Gretchen and Sutskever, Ilya},
  booktitle={Proceedings of the 38th International Conference on Machine Learning},
  editor={Meila, Marina and Zhang, Tong},
  series={Proceedings of Machine Learning Research},
  volume={139},
  pages={8748--8763},
  publisher={PMLR},
  month={18--24 Jul},
  year={2021},
  url={https://proceedings.mlr.press/v139/radford21a.html}
}

@inproceedings{desai2023hyperbolic,
  title={Hyperbolic image-text representations},
  author={Desai, Karan and Nickel, Maximilian and Rajpurohit, Tanmay and Johnson, Justin and Vedantam, Shanmukha Ramakrishna},
  booktitle={Proceedings of the 40th International Conference on Machine Learning},
  editor={Krause, Andreas and Brunskill, Emma and Cho, Kyunghyun and Engelhardt, Barbara and Sabato, Sivan and Scarlett, Jonathan},
  series={Proceedings of Machine Learning Research},
  volume={202},
  pages={7694--7731},
  publisher={PMLR},
  month={23--29 Jul},
  year={2023},
  url={https://proceedings.mlr.press/v202/desai23a.html}
}

@inproceedings{ramasinghe2024accept,
  title={Accept the modality gap: An exploration in the hyperbolic space},
  author={Ramasinghe, Sameera and Shevchenko, Violetta and Avraham, Gil and Thalaiyasingam, Ajanthan},
  booktitle={Proceedings of the IEEE/CVF Conference on Computer Vision and Pattern Recognition (CVPR)},
  pages={27263--27272},
  month={June},
  year={2024}
}

@inproceedings{pal2024compositional,
  title={Compositional Entailment Learning for Hyperbolic Vision-Language Models},
  author={Pal, Avik and van Spengler, Max and di Melendugno, Guido Maria D'Amely and Flaborea, Alessandro and Galasso, Fabio and Mettes, Pascal},
  booktitle={The Thirteenth International Conference on Learning Representations},
  year={2025},
  url={https://openreview.net/forum?id=3i13Gev2hV}
}

@inproceedings{kim2026uncha,
  title={Uncertainty-guided Compositional Alignment with Part-to-Whole Semantic Representativeness in Hyperbolic Vision-Language Models},
  author={Kim, Hayeon and Jang, Ji Ha and Kim, Junghun James and Chun, Se Young},
  booktitle={Proceedings of the IEEE/CVF Conference on Computer Vision and Pattern Recognition (CVPR)},
  pages={36861--36870},
  month={June},
  year={2026}
}

@inproceedings{yoshikawa2026phyclip,
  title={{PHyCLIP}: {$\ell_1$}-Product of Hyperbolic Factors Unifies Hierarchy and Compositionality in Vision-Language Representation Learning},
  author={Yoshikawa, Daiki and Matsubara, Takashi},
  booktitle={The Fourteenth International Conference on Learning Representations},
  year={2026},
  url={https://openreview.net/forum?id=I3Ct1eDmVI}
}

@misc{peng2023kosmos,
  title={Kosmos-2: Grounding multimodal large language models to the world},
  author={Peng, Zhiliang and Wang, Wenhui and Dong, Li and Hao, Yaru and Huang, Shaohan and Ma, Shuming and Wei, Furu},
  year={2023},
  url={https://arxiv.org/abs/2306.14824}
}

@inproceedings{alper2024hierarcaps,
  title={Emergent Visual-Semantic Hierarchies in Image-Text Representations},
  author={Alper, Morris and Averbuch-Elor, Hadar},
  booktitle={Computer Vision -- ECCV 2024},
  pages={220--238},
  publisher={Springer Nature Switzerland},
  year={2024},
  doi={10.1007/978-3-031-72943-0_13}
}

@inproceedings{lin2014microsoft,
  title={Microsoft {COCO}: Common Objects in Context},
  author={Lin, Tsung-Yi and Maire, Michael and Belongie, Serge and Hays, James and Perona, Pietro and Ramanan, Deva and Doll{\'a}r, Piotr and Zitnick, C. Lawrence},
  booktitle={Computer Vision -- ECCV 2014},
  pages={740--755},
  publisher={Springer International Publishing},
  year={2014},
  doi={10.1007/978-3-319-10602-1_48}
}

@inproceedings{plummer2015flickr30k,
  title={Flickr30k Entities: Collecting Region-to-Phrase Correspondences for Richer Image-to-Sentence Models},
  author={Plummer, Bryan A. and Wang, Liwei and Cervantes, Chris M. and Caicedo, Juan C. and Hockenmaier, Julia and Lazebnik, Svetlana},
  booktitle={Proceedings of the IEEE International Conference on Computer Vision (ICCV)},
  pages={2641--2649},
  year={2015}
}

@inproceedings{deng2009imagenet,
  title={{ImageNet}: A Large-Scale Hierarchical Image Database},
  author={Deng, Jia and Dong, Wei and Socher, Richard and Li, Li-Jia and Li, Kai and Fei-Fei, Li},
  booktitle={2009 IEEE Conference on Computer Vision and Pattern Recognition},
  pages={248--255},
  publisher={IEEE},
  year={2009},
  doi={10.1109/CVPR.2009.5206848}
}

@inproceedings{zohra2025betaclip,
  title={{b-CLIP}: Text-Conditioned Contrastive Learning for Multi-Granular Vision-Language Alignment},
  author={Zohra, Fatimah and Zhao, Chen and Itani, Hani and Ghanem, Bernard},
  booktitle={Proceedings of the IEEE/CVF Conference on Computer Vision and Pattern Recognition (CVPR)},
  pages={680--689},
  month={June},
  year={2026}
}

@inproceedings{jia2021scaling,
  title={Scaling up visual and vision-language representation learning with noisy text supervision},
  author={Jia, Chao and Yang, Yinfei and Xia, Ye and Chen, Yi-Ting and Parekh, Zarana and Pham, Hieu and Le, Quoc V. and Sung, Yun-Hsuan and Li, Zhen and Duerig, Tom},
  booktitle={Proceedings of the 38th International Conference on Machine Learning},
  editor={Meila, Marina and Zhang, Tong},
  series={Proceedings of Machine Learning Research},
  volume={139},
  pages={4904--4916},
  publisher={PMLR},
  month={18--24 Jul},
  year={2021},
  url={https://proceedings.mlr.press/v139/jia21b.html}
}

@inproceedings{zhai2022lit,
  title={{LiT}: Zero-Shot Transfer with Locked-image Text Tuning},
  author={Zhai, Xiaohua and Wang, Xiao and Mustafa, Basil and Steiner, Andreas and Keysers, Daniel and Kolesnikov, Alexander and Beyer, Lucas},
  booktitle={Proceedings of the IEEE/CVF Conference on Computer Vision and Pattern Recognition (CVPR)},
  pages={18102--18112},
  month={June},
  year={2022}
}

@inproceedings{zhai2023sigmoid,
  title={Sigmoid loss for language image pre-training},
  author={Zhai, Xiaohua and Mustafa, Basil and Kolesnikov, Alexander and Beyer, Lucas},
  booktitle={Proceedings of the IEEE/CVF International Conference on Computer Vision (ICCV)},
  pages={11975--11986},
  month={October},
  year={2023}
}

@inproceedings{zhong2022regionclip,
  title={{RegionCLIP}: Region-Based Language-Image Pretraining},
  author={Zhong, Yiwu and Yang, Jianwei and Zhang, Pengchuan and Li, Chunyuan and Codella, Noel and Li, Liunian Harold and Zhou, Luowei and Dai, Xiyang and Yuan, Lu and Li, Yin and Gao, Jianfeng},
  booktitle={Proceedings of the IEEE/CVF Conference on Computer Vision and Pattern Recognition (CVPR)},
  pages={16793--16803},
  month={June},
  year={2022}
}

@inproceedings{li2022grounded,
  title={Grounded Language-Image Pre-training},
  author={Li, Liunian Harold and Zhang, Pengchuan and Zhang, Haotian and Yang, Jianwei and Li, Chunyuan and Zhong, Yiwu and Wang, Lijuan and Yuan, Lu and Zhang, Lei and Hwang, Jenq-Neng and Chang, Kai-Wei and Gao, Jianfeng},
  booktitle={Proceedings of the IEEE/CVF Conference on Computer Vision and Pattern Recognition (CVPR)},
  pages={10965--10975},
  month={June},
  year={2022}
}

@inproceedings{li2021align,
  title={Align before fuse: Vision and language representation learning with momentum distillation},
  author={Li, Junnan and Selvaraju, Ramprasaath and Gotmare, Akhilesh and Joty, Shafiq and Xiong, Caiming and Hoi, Steven Chu Hong},
  booktitle={Advances in Neural Information Processing Systems},
  editor={Ranzato, Marc'Aurelio and Beygelzimer, Alina and Dauphin, Yann and Liang, Percy S. and Vaughan, Jennifer Wortman},
  volume={34},
  pages={9694--9705},
  publisher={Curran Associates, Inc.},
  year={2021},
  url={https://proceedings.neurips.cc/paper_files/paper/2021/file/505259756244493872b7709a8a01b536-Paper.pdf}
}

@inproceedings{li2022blip,
  title={{BLIP}: Bootstrapping Language-Image Pre-training for Unified Vision-Language Understanding and Generation},
  author={Li, Junnan and Li, Dongxu and Xiong, Caiming and Hoi, Steven},
  booktitle={Proceedings of the 39th International Conference on Machine Learning},
  editor={Chaudhuri, Kamalika and Jegelka, Stefanie and Song, Le and Szepesvari, Csaba and Niu, Gang and Sabato, Sivan},
  series={Proceedings of Machine Learning Research},
  volume={162},
  pages={12888--12900},
  publisher={PMLR},
  month={17--23 Jul},
  year={2022},
  url={https://proceedings.mlr.press/v162/li22n.html}
}

@inproceedings{li2023blip,
  title={{BLIP}-2: Bootstrapping Language-Image Pre-training with Frozen Image Encoders and Large Language Models},
  author={Li, Junnan and Li, Dongxu and Savarese, Silvio and Hoi, Steven},
  booktitle={Proceedings of the 40th International Conference on Machine Learning},
  editor={Krause, Andreas and Brunskill, Emma and Cho, Kyunghyun and Engelhardt, Barbara and Sabato, Sivan and Scarlett, Jonathan},
  series={Proceedings of Machine Learning Research},
  volume={202},
  pages={19730--19742},
  publisher={PMLR},
  month={23--29 Jul},
  year={2023},
  url={https://proceedings.mlr.press/v202/li23q.html}
}

@inproceedings{ganea2018hyperbolic,
  title={Hyperbolic entailment cones for learning hierarchical embeddings},
  author={Ganea, Octavian and B{\'e}cigneul, Gary and Hofmann, Thomas},
  booktitle={Proceedings of the 35th International Conference on Machine Learning},
  editor={Dy, Jennifer and Krause, Andreas},
  series={Proceedings of Machine Learning Research},
  volume={80},
  pages={1646--1655},
  publisher={PMLR},
  month={10--15 Jul},
  year={2018},
  url={https://proceedings.mlr.press/v80/ganea18a.html}
}

@inproceedings{nickel2017poincare,
  title={Poincar{\'e} embeddings for learning hierarchical representations},
  author={Nickel, Maximillian and Kiela, Douwe},
  booktitle={Advances in Neural Information Processing Systems},
  editor={Guyon, Isabelle and Von Luxburg, Ulrike and Bengio, Samy and Wallach, Hanna and Fergus, Rob and Vishwanathan, S. and Garnett, Roman},
  volume={30},
  publisher={Curran Associates, Inc.},
  year={2017},
  url={https://proceedings.neurips.cc/paper_files/paper/2017/file/59dfa2df42d9e3d41f5b02bfc32229dd-Paper.pdf}
}

@inproceedings{law2019lorentzian,
  title={Lorentzian distance learning for hyperbolic representations},
  author={Law, Marc and Liao, Renjie and Snell, Jake and Zemel, Richard},
  booktitle={Proceedings of the 36th International Conference on Machine Learning},
  editor={Chaudhuri, Kamalika and Salakhutdinov, Ruslan},
  series={Proceedings of Machine Learning Research},
  volume={97},
  pages={3672--3681},
  publisher={PMLR},
  month={09--15 Jun},
  year={2019},
  url={https://proceedings.mlr.press/v97/law19a.html}
}

@inproceedings{nickel2018learning,
  title={Learning continuous hierarchies in the {Lorentz} model of hyperbolic geometry},
  author={Nickel, Maximillian and Kiela, Douwe},
  booktitle={Proceedings of the 35th International Conference on Machine Learning},
  editor={Dy, Jennifer and Krause, Andreas},
  series={Proceedings of Machine Learning Research},
  volume={80},
  pages={3779--3788},
  publisher={PMLR},
  month={10--15 Jul},
  year={2018},
  url={https://proceedings.mlr.press/v80/nickel18a.html}
}

@inproceedings{liu2024grounding,
  title={Grounding {DINO}: Marrying {DINO} with Grounded Pre-Training for Open-Set Object Detection},
  author={Liu, Shilong and Zeng, Zhaoyang and Ren, Tianhe and Li, Feng and Zhang, Hao and Yang, Jie and Jiang, Qing and Li, Chunyuan and Yang, Jianwei and Su, Hang and Zhu, Jun and Zhang, Lei},
  booktitle={Computer Vision -- ECCV 2024},
  pages={38--55},
  publisher={Springer Nature Switzerland},
  year={2025},
  doi={10.1007/978-3-031-72970-6_3}
}

@misc{he2025position,
  title={Position: Beyond Euclidean -- Foundation Models Should Embrace Non-Euclidean Geometries},
  author={He, Neil and Liu, Jiahong and Zhang, Buze and Bui, Ngoc and Maatouk, Ali and Yang, Menglin and King, Irwin and Weber, Melanie and Ying, Rex},
  year={2025},
  url={https://arxiv.org/abs/2504.08896}
}

\end{document}